\documentclass[10pt,letterpaper]{article}
 
\usepackage[top=5cm, bottom=5cm, left=3.8cm, right=3.8cm]{geometry}
 
\usepackage[T1]{fontenc}
\usepackage[utf8]{inputenc}
\usepackage{mathptmx}     % Times New Roman as main font
\usepackage{courier}      % Courier for typewriter / algorithm font
 
\usepackage{indentfirst}
\usepackage{booktabs}
\usepackage{graphicx}
\usepackage{amsmath}
\usepackage{amssymb}
\usepackage{enumitem}
\usepackage{hyperref}

\hypersetup{
  colorlinks=false,
  pdfborder={0 0 1},
  citebordercolor={1 0 0},
  linkbordercolor={1 1 1},
  urlbordercolor={1 1 1},
  filebordercolor={1 1 1}
}
 
\usepackage[font=small,labelfont=bf]{caption}
\usepackage{titling}
\pretitle{\begin{center}\fontsize{16pt}{19pt}\selectfont\bfseries}
\posttitle{\par\end{center}\vskip 1ex}
 
\renewenvironment{abstract}
{\begin{quote}
 \noindent {\small \bfseries \abstractname.} \small
}
{\medskip\noindent
 \end{quote}
}
 
\newenvironment{keywords}
{\begin{quote}
 \noindent {\small \bfseries Keywords: } \small
}
{\medskip\noindent
 \end{quote}
}
 
\usepackage{titlesec}
\titleformat*{\section}{\large\bfseries}
\titleformat*{\subsection}{\normalsize\bfseries}
\titleformat*{\subsubsection}{\normalsize\bfseries}
\titleformat*{\paragraph}{\normalsize\bfseries}
\titleformat*{\subparagraph}{\normalsize\bfseries}

\usepackage[plain, noline]{algorithm2e}
\SetAlFnt{\fontfamily{pcr}\selectfont}
\SetAlCapSkip{2ex}
\IncMargin{-1em}
\SetKwInput{KwIn}{Input}
\SetKwInput{KwOut}{Output}
\SetKwIF{If}{ElseIf}{Else}{if}{then}{else\ if}{else}{end\ if}
\SetKwFor{For}{for}{do}{end\ for}
\SetKw{KwTo}{until}
\SetKwRepeat{Repeat}{repeat}{until}
\SetKwFor{While}{while}{do}{end\ while}
 
\usepackage[hang,flushmargin]{footmisc}
 
\usepackage{bibspacing}
\begin{document}
 
\title{Swin Meets EfficientNet: Lightweight Architectures for GAN-Based Face Forensics}
 
\author{%
\parbox{\dimexpr\textwidth-2\tabcolsep\relax}{%
\centering
\makebox[\linewidth][c]{\normalsize Sejuti Basu$^{1}$, Ashima Sood$^{2}$, Vijay Kumar$^{3}$, Sahil Sharma$^{2}$}\\
\makebox[\linewidth][c]{\small $\lbrace$basusejuti@gmail.com, sood-a1@ulster.ac.uk, vijaykumarchahar@gmail.com, s.sharma@ulster.ac.uk$\rbrace$}\\[4pt]
\makebox[\linewidth][c]{\small $^{1}$International Institute of Information Technology, Bangalore, India}\\
\makebox[\linewidth][c]{\small $^{2}$School of Computing, Engineering, and Intelligent Systems, Ulster University, Londonderry, United Kingdom}\\
\makebox[\linewidth][c]{\small $^{3}$Department of Information Technology, Dr B R Ambedkar National Institute of Technology Jalandhar, Punjab, India}
}%
}
 
\date{}
 
\pagestyle{empty}
 
\maketitle
\thispagestyle{empty}
 
\vspace{-3.75em}

\begin{abstract}
Modern generative models, such as GANs, diffusion architectures, and autoregressive systems, now produce facial images that are nearly indistinguishable from authentic photographs. This capability makes detecting forged images increasingly difficult, raising serious concerns about identity theft, fraud, and misinformation campaigns. Our research focuses specifically on GAN-generated synthetic faces, which underpin many face-centric deepfakes, and investigates efficient detection approaches using image analysis alone. Existing detection systems rely heavily on either convolutional neural networks (CNNs) or global vision transformers. While CNNs excel at identifying texture-based local features, they struggle with broader contextual understanding. Traditional Vision Transformer (ViT) models can capture long-range structures effectively, but demand substantial computational resources. Our work explores Swin-Transformer-based architectures across three implementations: a compact Swin Transformer trained from the ground up, ImageNet-1K pre-trained Swin-Tiny and Swin-Small models adapted for binary classification, and a novel hybrid combining EfficientNet-B0's convolutional processing with a Swin Transformer backend. We evaluated all models using the 140K Real and Fake Faces dataset, which includes StyleGAN-generated fake faces alongside authentic images from Flickr and DFDC, with balanced splits for training, validation, and testing. The EfficientNetB0+Swin hybrid achieved 99\% accuracy and a 99.44\% recall on 5,000 test images, outperforming both pure Swin variants and a previous CNN-only baseline on this dataset. Our results suggest that combining hierarchical CNN features with shifted-window self-attention provides an efficient and computationally lightweight method for detecting GAN-generated synthetic faces.
\end{abstract}

% \begin{keywords}
% Deepfake, Transformers, EfficientNet, Swin-Transformer
% \end{keywords}

\section{Introduction}
\label{sec:intro}

Facial images are commonly utilised as biometric identifiers. Mobile devices, border control systems, financial institutions, and social media platforms collect such data, which is increasingly repurposed for training large-scale recognition and generative models. At the same time, concerns about biometric leakage and unauthorised reuse of facial data without proper consent continue to grow. These tensions have made data synthesis an important research direction in biometrics. Synthetic faces could help facilitate model development and evaluation while reducing dependence on identifiable subjects and raw operational logs. Regulators and international organisations stress the need for stronger protections against harmful uses of synthetic media, including identity theft, financial fraud, political misinformation, and reputational damage.\footnote{\href{https://www.unesco.org/en/articles/deepfakes-and-crisis-knowing}{https://www.unesco.org/en/articles/deepfakes-and-crisis-knowing}}

In this environment, high-fidelity face synthesis has become widespread. Beyond classical Generative Adversarial Networks (GANs)~\cite{goodfellow2014generative}, modern diffusion and autoregressive models can generate photorealistic human faces and perform effective face editing~\cite{ho2020denoising,tian2024visual}. Our study examines synthetic faces produced by GANs, specifically architectures similar to StyleGAN~\cite{karras2019style}. These models play a central role in current biometric practices, being used to construct large-scale synthetic face datasets, generate identities with controllable attributes, and support attack pipelines requiring rapid and cost-effective identity fabrication. From a forensic standpoint, GAN-generated faces display unique artifact patterns and distributional biases distinct from those in diffusion models. Understanding how detector architectures respond to GAN-specific signatures is crucial for improving biometric system security and for clarifying the limitations of synthetic face datasets used in training and evaluation.

Deepfake detection is typically framed as a supervised binary classification problem for images or videos, though effective modeling requires more than simple label assignment. Detectors must respond to detailed local evidence like texture inconsistencies, blending artifacts, and abnormal noise statistics, while also accounting for global facial structure and cross-region consistency~\cite{rossler2019faceforensics++,abbas2024unmasking}.Convolutional neural networks (CNNs) are frequently used here because they leverage local correlations and perform well across diverse benchmarks. Backbones like VGG, ResNet, Xception, and EfficientNet focus on local patterns. Although their effective receptive field expands with depth, they model global context only indirectly~\cite{gupta2023comprehensive}. Vision Transformers (ViTs)~\cite{dosovitskiy2021an} take a different approach by applying global self-attention across all patches. This enables robust long-range modeling but comes with higher memory and computational costs, plus greater reliance on extensive pretraining. The Swin Transformer~\cite{liu2021swin} offers a middle ground by combining shifted window self-attention with a hierarchical architecture, balancing local detail, global structure, and computational efficiency. Our paper investigates how architectural choices affect the detection of GAN-generated synthetic faces using Swin-based architectures and a hybrid CNN-Swin design. This approach computes self-attention within non-overlapping windows, alternating with shifted windows, yielding a hierarchical representation with three useful properties for deepfake detection: (i) a multi-scale feature pyramid capturing both fine-grained and semantic information, (ii) self-attention with computational cost scaling linearly with image size, and (iii) cross-window interactions approximating global context without requiring full global attention. Swin is lighter and simpler to implement than ViT-style global transformers, while offering better contextual modeling than pure CNNs.

The primary concern is \emph{how far can Swin-based architectures be advanced for GAN-based synthetic face detection, and how do they perform in comparison to robust CNN baselines?} We analyse Swin both independently and in conjunction with an EfficientNet backbone, utilising StyleGAN-generated faces from the 140K Real and Fake Faces dataset as a controlled testbed. This research emphasises GAN-based images, yet the architectural insights are applicable to other generative models.

Our contributions are as follows:
\begin{itemize}
    \item This study systematically investigates Swin-Transformer architectures for deepfake image detection, analysing the impact of design choices, including patch size, window size, and training regime (from scratch versus transfer learning), on performance with GAN-generated synthetic faces.
    \item This study presents a hybrid architecture that merges EfficientNet-B0 with a Swin-Transformer head. This integration facilitates convolutional feature extraction alongside shifted-window self-attention, enabling the joint modelling of local artefacts and global facial context, while maintaining a significantly lighter profile compared to a standard ViT.
    \item This study assesses three Swin-based variants: a task-specific Swin trained from scratch, ImageNet-pretrained Swin-Tiny and Swin-Small, and the proposed hybrid EfficientNet+Swin, using the 140K Real and Fake Faces dataset. These variants are compared to an EfficientNet-based baseline. The hybrid model demonstrates 99\% accuracy and a 99.44\% recall on a test set of 5,000 images, significantly surpassing the performance of both solely Swin-based and CNN-based detectors on this GAN-centric benchmark.
\end{itemize}

The remainder of the paper is organised as follows. Section~\ref{sec:related} reviews related work on deepfake detection with CNNs and transformers. Section~\ref{sec:method} introduces our Swin-based and hybrid architectures. Section~\ref{sec:experiments} presents the experimental setup and results. Section~\ref{sec:conclusion} concludes with a discussion of limitations and directions for extending this work beyond GAN-based synthetic faces.

\section{Related Work}
\label{sec:related}

Initial research focused on manually designed cues and traditional image forensics while contemporary approaches are primarily characterised by deep learning models that function directly on pixel data. This section outlines three primary categories: CNN-based detectors, transformer and hybrid architectures, and Swin-Transformer-based approaches. Our work is situated at the convergence of robustness and computational efficiency with a clear focus on resource-aware model design.

\textbf{CNN-based detectors:} CNNs remain the dominant framework for deepfake detection because of their strong locality bias and favorable accuracy-efficiency tradeoff. Architectures like VGG, ResNet, and Xception have been adapted to emphasize spatial artifacts and abnormal noise statistics in manipulated facial images. Chang et al. use SRM filtering to highlight noise residuals before feeding them to a VGG-16 classifier~\cite{chang2020deepfake}. Bondi et al. show that carefully designed augmentations and loss functions substantially improve CNN performance on DFDC~\cite{bondi2020training}. EfficientNet variants have been explored as more parameter-efficient backbones for both curated and in-the-wild datasets~\cite{deng2022deepfake}. Beyond standard CNNs, several studies incorporate frequency-domain or spectral information~\cite{li2024freqblender}. Earlier work~\cite{kumar2021deepfake} combines Error Level Analysis (ELA) with a CNN backbone on the 140K Real and Fake Faces dataset, creating a competitive all-convolutional baseline.

\textbf{Transformer and hybrid detectors}: Vision transformers have been implemented in deepfake detection to effectively address the deficiency of global context. ViT-style models utilise self-attention across all image patches, enabling robust long-range modelling that effectively captures inconsistencies in global facial layout, illumination, and geometry. Coccomini et al. integrate an EfficientNet backbone with a ViT branch through cross-attention, effectively merging local and global cues for the purpose of video forgery detection~\cite{coccomini2022combining}. 
A range of hybrid CNN-transformer architectures has been developed, wherein a convolutional stage extracts low-level features while a transformer stage analyzes higher-level context. Research on deepfake detection consistently indicates that hybrid models represent a promising avenue, as they combine the inductive biases and efficiency of convolutional neural networks (CNNs) with the modelling flexibility of transformers. Global transformers, such as ViT-like models, exhibit a quadratic cost relative to the number of patches, rendering them less attractive for practitioners constrained by practical computational and energy limitations.

\textbf{Swin-Transformer-based approaches}: Swin-Transformer models~\cite{liu2021swin} mitigate scalability challenges by confining self-attention to non-overlapping windows and employing shifted windows to facilitate interactions across windows. This results in a hierarchical feature pyramid characterised by linear complexity relative to image size. Swin has been utilised in various deepfake-related applications, such as serving as a backbone for thumbnail-based video detectors~\cite{xu2023tall} or as a component in more intricate multi-branch architectures~\cite{ilyas2023avfakenet}. Recent studies have investigated Swin-based detectors aimed at enhancing cross-manipulation robustness and cross-dataset generalisation, demonstrating that local-global reasoning within a hierarchical transformer can improve synthetic face detection~\cite{gong2024swin}.
Many current Swin-based methods either operate at high input resolutions with substantial backbone sizes or are integrated within larger frameworks that incur a high overall computational cost.

\textbf{Positioning and scope of this work}: This study concentrates on GAN-based synthetic faces and architectures that can be trained and deployed with limited computational resources. Swin-Transformer variants represent an intermediary between CNNs and global ViTs, providing enhanced contextual reasoning compared to CNN-only baselines while maintaining a significantly lower computational cost than traditional ViTs. Swin-based configurations (Swin Base, Swin Transfer, and Hybrid EfficientNet+Swin) aim to provide a compact, reproducible framework for practitioners who require practical, resource-efficient detectors rather than a singular optimally large model.

Prior research indicates that deep and complex architectures achieve high performance on various deepfake benchmarks. Our contribution demonstrates that well-designed Swin-based and hybrid models can be comparable or exceed robust CNN baselines on a GAN-based dataset without the need for large-scale training or advanced hardware. This establishes our method as a robust, resource-efficient advancement in scalable synthetic face forensics.

\section{Method}
\label{sec:method}

The objective is to develop an image-only detector capable of differentiating between authentic faces and GAN-generated synthetic faces while adhering to practical computational limitations. This section first formalises the problem and the design principles that inform our choices, followed by a description of the instantiation of Swin-Transformer backbones and their integration with EfficientNet in our hybrid architecture.

\subsection{Problem formulation and design principles}
\label{subsec:problem}
Let $\mathbf{I} \in \mathbb{R}^{H \times W \times 3}$ represent a cropped RGB face image, and let $y \in \{0,1\}$ denote its label, where $y = 1$ indicates a GAN-generated (synthetic) face and $y = 0$ indicates a real face. A detector with parameters $\theta$ assigns a score $p_\theta(\mathbf{I}) \in [0,1]$ to input $\mathbf{I}$, interpreted as the probability that $\mathbf{I}$ is synthetic.

We train all models as binary classifiers using standard binary cross-entropy loss on a training set $\mathcal{D}$:
\begin{equation}
    \mathcal{L}(\theta) 
    = - \frac{1}{|\mathcal{D}|} 
    \sum_{(\mathbf{I},\,y) \in \mathcal{D}} 
    \big[\, y \log p_\theta(\mathbf{I}) 
    + (1-y) \log \big(1 - p_\theta(\mathbf{I})\big) \big].
    \label{eq:bce_loss}
\end{equation}

The complete 140K Real and Fake Faces dataset contains 140,000 images, but we work with a balanced subset: 25,000 training images, 5,000 validation images, and 5,000 test images for this binary real-fake classification task (see Section~\ref{sec:experiments}). The image quality and diversity in this subset are sufficient for building a robust discriminator. Using all 140,000 images would yield only marginal improvements while substantially increasing computational and environmental costs. This choice reflects our core design philosophy.

We establish three guiding principles for our design. First, we focus on a GAN framework within an image-only setting: the detector trains on StyleGAN-generated synthetic faces and operates exclusively with still images, not using temporal information or extra metadata. Second, we aim to build architectures that are both context-aware and efficient, going beyond local CNN filters by incorporating long-range interactions while keeping memory footprint and FLOP count much lower than global vision transformers. Third, rather than proposing a single large model optimized for one benchmark, we present a compact set of Swin-based configurations that can be trained on standard hardware and adapted to other synthetic face datasets with minimal tuning, providing a resource-efficient framework for practical deployment.

The choice of Swin-Transformer backbones and their integration with EfficientNet-B0 follows from these principles, which we discuss next.
\subsection{Swin-Transformer backbone for face forensics}
\label{subsec:swin_backbone}

The Swin Transformer~\cite{liu2021swin} offers a hierarchical vision backbone that substitutes global self-attention with local, window-based attention utilising shifted windows. This paper summarises the components utilised and their specific applications in face forgery detection.

Given an input image $\mathbf{I}$, Swin initially divides it into non-overlapping patches of size $P \times P$ and subsequently linearly projects each patch into a $C$-dimensional token. Tokens undergo a series of processing stages. The stages comprise:

\begin{enumerate}[leftmargin=*]
    \item \emph{Window Multi-Head Self-Attention (W-MSA)}, which computes self-attention independently within fixed-size windows (e.g., $M \times M$ patches).
    \item \emph{Shifted Window Multi-Head Self-Attention (SW-MSA)}, which applies the same operation after cyclically shifting the window partition, enabling cross-window interactions.
    \item \emph{Feed-forward MLP blocks} with GELU activations and residual connections.
    \item \emph{Patch merging} layers between stages, which halve the spatial resolution and increase the channel dimension to build a multi-scale representation.
\end{enumerate}

The above design produces a feature pyramid that effectively captures fine-grained local structures in the initial stages and more semantic, global information in the deeper stages. In the context of face forensics, this approach is advantageous: early layers can effectively model skin texture, boundary artefacts, and local inconsistencies, whereas deeper layers can assess global facial coherence (e.g., consistency among the eyes, mouth, and background) at a cost that scales approximately linearly with image size.
In all Swin-based models, the final token sequence undergoes global average pooling and is subsequently mapped to a scalar logit, which is processed through a sigmoid function to yield $p_\theta(\mathbf{I})$.

\subsection{Model variants}
\label{subsec:model_variants}

\begin{figure}[!htb]
    \centering
   \includegraphics[width=0.9\textwidth,height=0.41\textheight]{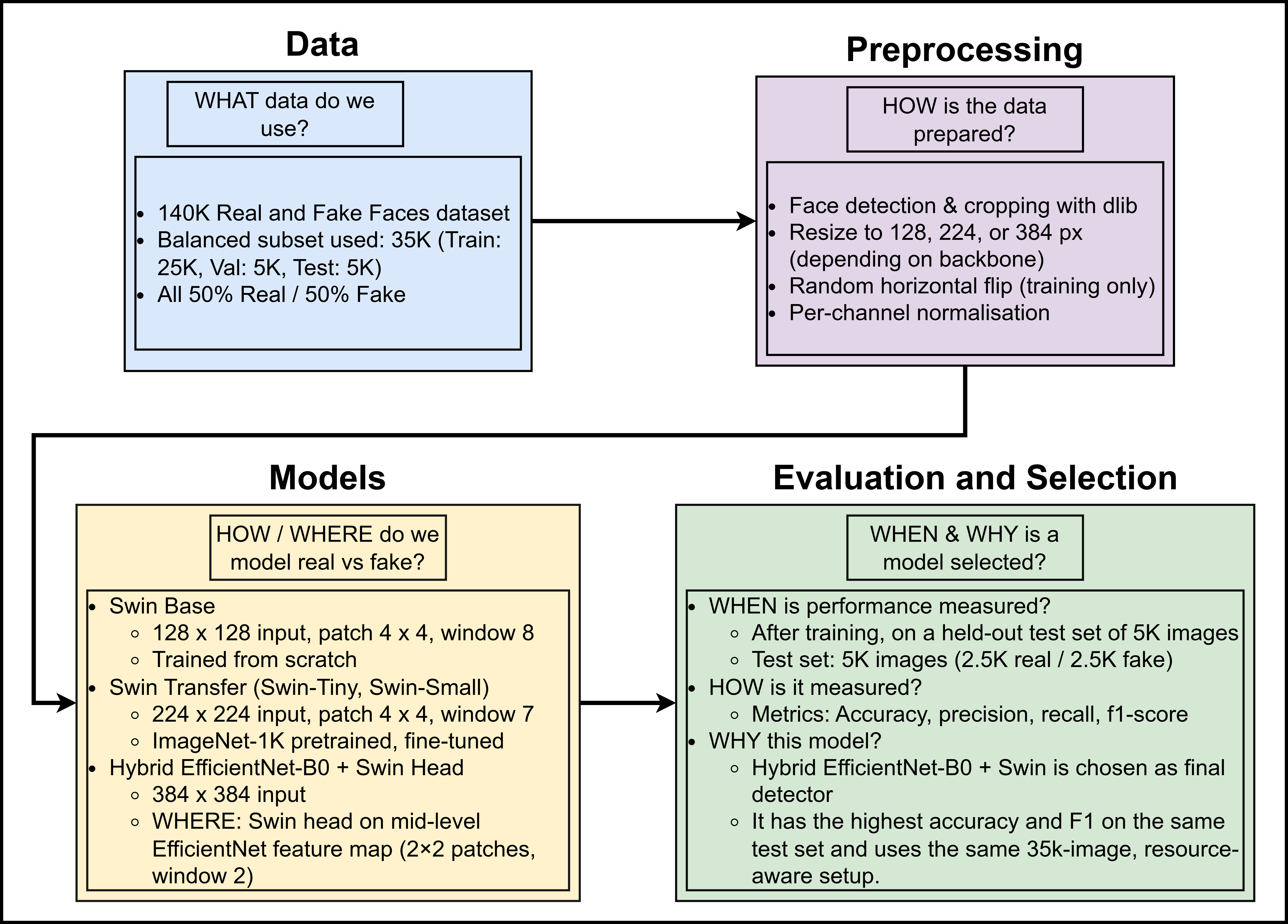} 
   \caption{Proposed deepfake detection pipeline, organised according to the \textit{what/how/where/when/why} framework.}

   \label{fig:pipeline}
\end{figure}

We instantiate the above backbone in three ways, shown schematically in Figure~\ref{fig:pipeline}. The variants differ in how much prior knowledge they exploit (from scratch vs.\ pretraining) and how they combine convolutional and transformer-based reasoning.

\subsubsection{Swin Base: task-specific Swin from scratch}
\label{subsubsec:swin_base}

The Swin Base model is a compact Swin-Transformer designed specifically for GAN-based face detection and trained from scratch on a subset of our dataset. We employ a low input resolution (e.g., $128 \times 128$) and small window sizes to maintain a reduced parameter count and FLOPs, while still leveraging hierarchical attention.
This variant fulfils two functions. Initially, it serves as a baseline for Swin that does not utilise ImageNet pretraining, enabling an assessment of Swin's capability to develop effective artefact detectors from GAN data. Additionally, it serves as a robust benchmark for assessing the comparative advantages of transfer learning and hybridisation.

\subsubsection{Swin Transfer: ImageNet-pretrained Swin-Tiny and Swin-Small}
\label{subsubsec:swin_transfer}

The Swin Transfer configuration employs pretraining on ImageNet-1K. We utilise two standard backbones, Swin-Tiny\footnote{https://huggingface.co/microsoft/swin-tiny-patch4-window7-224} and Swin-Small,\footnote{https://huggingface.co/microsoft/swin-small-patch4-window7-224}, resizing all input images to $224 \times 224$. The original ImageNet classification head is substituted with a single sigmoid unit, and all layers, including the Swin backbone, undergo fine-tuning on the deepfake dataset.

 Analysing Swin-Tiny and Swin-Small facilitates an examination of the accuracy-efficiency trade-off present in the Swin family. Swin-Small possesses an increased number of layers and broader channels, resulting in elevated computational costs. Our experiments in Section~\ref{sec:experiments} demonstrate that, for a binary GAN-based task, the smaller Swin-Tiny outperforms its larger counterpart, thereby supporting the use of compact backbones in this context.

\subsubsection{Hybrid EfficientNet and Swin integration}
\label{subsubsec:hybrid_swin}

The primary architecture employed is a hybrid model that combines EfficientNet-B0 with a Swin-Transformer head. The design is inspired by the observation that convolutional networks excel at capturing fine-grained local artefacts, while Swin-Transformers are more adept at modelling broader facial context. The aim is to utilise both types of cues while maintaining a lightweight overall model.
Each incoming face is processed at a resolution of $384 \times 384$ using an EfficientNet-B0 backbone pre-trained with ImageNet-1K weights. We extract two complementary representations from this backbone: a mid-level feature map from an intermediate block\footnote{\texttt{block6a\_expand\_activation}}, which retains spatial detail and rich texture information, and a final feature map from the last convolutional layer, encoding higher-level semantics in a more compressed spatial form. The mid-level feature map is divided into small patches (e.g., $2 \times 2$), linearly embedded, and processed through a shallow Swin stack with a limited window size and number of heads. The Swin head enhances convolutional features by modelling interactions within and between facial regions at a lower spatial resolution. The final EfficientNet feature map undergoes global average pooling to generate a compact CNN descriptor.

The outputs from these two branches are subsequently integrated in a straightforward fusion stage. Global average pooling is applied to the Swin output to generate a single vector, which is subsequently concatenated with the pooled EfficientNet descriptor. The joint representation is subsequently processed by a fully connected layer utilising sigmoid activation to yield $p_\theta(\mathbf{I})$. This hybrid design offers three practical advantages for GAN-based face forensics. Initially, it integrates complementary information: EfficientNet-B0 focuses on local artefacts and noise patterns, whereas the Swin head incorporates global context and cross-region consistency. Secondly, Swin processes a mid-level feature map instead of the raw input image, resulting in a reduced number of tokens and maintaining a low cost of self-attention. Third, we initialise EfficientNet-B0 using ImageNet-1K weights and train the Swin head on its features, enhancing data efficiency and aiding the model in generalising from a relatively small training set.

Section~\ref{sec:experiments} demonstrates that the EfficientNet+Swin hybrid significantly surpasses both the pure Swin variants and the prior CNN-only baseline on the same dataset, while maintaining a compact form suitable for deployment within standard computational constraints.

\section{Experiments and Discussion}
\label{sec:experiments}

\subsection{Dataset}
\label{subsec:dataset}

All models are evaluated using the 140K Real and Fake Faces dataset\footnote{\href{https://www.kaggle.com/datasets/xhlulu/140k-real-and-fake-faces}{https://www.kaggle.com/datasets/xhlulu/140k-real-and-fake-faces}}, comprising 70,000 authentic faces and 70,000 faces generated by StyleGAN. The real images are obtained from Flickr and DeepFake Detection Challenge (DFDC)~\cite{dolhansky2020deepfake}, encompassing a diverse array of ages, demographics, and accessories. In accordance with previous research~\cite{kumar2021deepfake}, faces are detected and cropped utilising \texttt{dlib}~\cite{king2009dlib}, subsequently resized to dimensions of $128$, $224$, or $384$ pixels based on the selected backbone. Random horizontal flipping and per-channel normalisation are applied during training.
In accordance with our resource-aware design, we utilise a balanced subset of the dataset for a binary real-fake classification task. We utilise 25,000 images for training, 5,000 for validation, and 5,000 for testing. Each subset comprises 50\% real and 50\% synthetic faces, resulting in 12,500 real and 12,500 synthetic images for training, and 2,500 real and 2,500 synthetic images for both validation and testing. The image quality and diversity within this subset are adequate for developing a robust discriminator, while utilising the complete set of 140k images would yield diminishing returns at a significantly increased computational expense.

\subsection{Implementation details}
\label{subsec:impl}

All models utilise the same balanced splits and are trained with the Adam optimiser. The Swin Base model functions with $128 \times 128$ inputs, utilising a patch size of $4 \times 4$ and a window size of 8. It is trained from scratch over 200 epochs, employing a learning rate of $2 \times 10^{-5}$ and a batch size of 32. Swin-Tiny and Swin-Small utilise $224 \times 224$ inputs, employing a patch size of 4 and a window size of 7. Both models are fine-tuned from ImageNet-1K pretraining, with a learning rate of $5 \times 10^{-4}$, batch sizes of 128 and 64, and training durations of 100 and 150 epochs, respectively. 

The hybrid EfficientNet+Swin model handles inputs of size $384 \times 384$. The Swin head is utilised on mid-level EfficientNet feature maps through the application of $2 \times 2$ patches and a window size of 2, employing 8 attention heads. The model undergoes training for 20 epochs, utilising a learning rate of 0.5 and a batch size of 8. Figure \ref{fig:Results}(a,b) illustrates the training dynamics. The model converges quickly, with the training loss decreasing from 0.97 to 0.20 in the first ten epochs, while the validation loss remains around 0.24 after nine epochs. Additionally, we can see that accuracy reaches 97\% by 11 epochs, compared to 99.8\% peak validation accuracy at epoch 17. Results are reported on the held-out 5,000-image test set unless stated otherwise.

\subsection{Quantitative results}
\label{subsec:results_quant}

Table~\ref{tab:comparison_sota} reports precision, recall, F1-score and accuracy on the test set, and compares our Swin-based architectures with the EfficientNet-based ELA+CNN baseline of Kumar et al. ~\cite{kumar2021deepfake}. The Swin Base model, trained from scratch, demonstrates performance comparable to the baseline, suggesting that a compact Swin configuration can rival a meticulously designed CNN approach on this GAN-based dataset without the need for external pretraining. The introduction of ImageNet-1K pretraining using Swin-Tiny results in a notable enhancement, with accuracy increasing from 88.50\% (Swin Base) to 91.13\%, and the F1-score increasing from 88.00\% to 91.14\%.

\begin{figure}[!htb]
    \centering
   \includegraphics[width=0.99\textwidth,height=0.42\textheight]{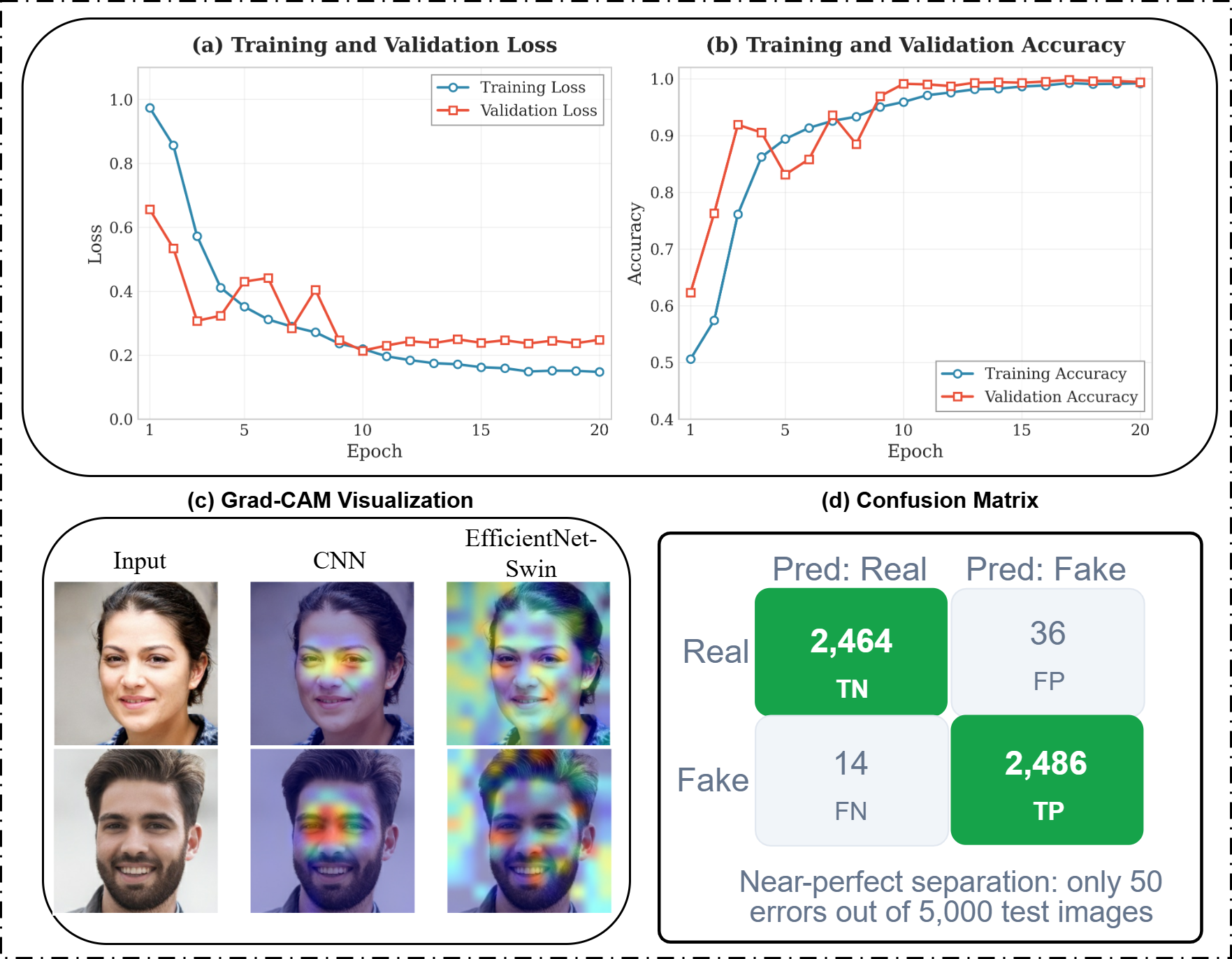}
   \caption{Performance over proposed EfficientNet-Swin hybrid Transformer model: (a) loss plots, (b) accuracy plots (peak: 99.8\%), (c) Grad-CAM visualizations comparing CNN and hybrid model attention regions, and (d) confusion matrix on 5,000 test images (recall: 99.44\%).}
   \label{fig:Results}
\end{figure}

Swin-Small exhibits lower performance than Swin-Tiny, despite its greater capacity and extended training duration, achieving an accuracy of 89.14\% and an F1-score of 89.13\%. This indicates that merely increasing model size does not inherently lead to improved performance in this binary GAN-based task, thereby supporting the argument for compact backbones.

The hybrid EfficientNet+Swin architecture achieves superior performance, with 98.57\% precision, 99.44\% recall, a 99\% F1-score, and 99\% accuracy (see Figure \ref{fig:Results}(d)). Compared to the ELA+CNN baseline, this represents an increase of over 11 percentage points in both accuracy and F1 Score, achieved with a relatively small model trained on a subset of the complete dataset. The findings substantiate our primary assertion that integrating convolutional features with shifted-window self-attention is notably effective for GAN-based synthetic face detection within realistic computational constraints.

\begin{table}[t]
    \centering
    \small
    \caption{Evaluation results on the 140K Real and Fake Faces test set (5{,}000 images).}
    \label{tab:comparison_sota}
    \begin{tabular}{p{2.5cm}p{1cm}p{1.5cm}p{1.5cm}p{1.5cm}p{1.5cm}}
        \hline
        \textbf{Model} & \textbf{Epochs} & \textbf{Precision} & \textbf{Recall} & \textbf{F1-score} & \textbf{Accuracy} \\
        \hline
        ELA + CNN~\cite{kumar2021deepfake}          & 20  & 86.76 & 88.40 & 87.57 & 87.46 \\
        \hline
        Swin Base (from scratch)           & 200 & 88.00 & 87.00 & 88.00 & 88.50 \\
        \hline
        Swin-Tiny (transfer learning)      & 100 & 88.00 & 94.00 & 91.14 & 91.13 \\
        \hline
        Swin-Small (transfer learning)     & 150 & 87.00 & 91.00 & 89.13 & 89.14 \\
        \hline
        Hybrid EfficientNet+Swin           & 20  & 98.57 & 99.44 & 99.00 & 99.00 \\
        \hline
    \end{tabular}
\end{table}

\subsection{Qualitative analysis and limitations}
\label{subsec:results_qual}

Grad-CAM visualisations for the hybrid model indicate attention to both local artefacts, including skin texture irregularities and boundary inconsistencies, as well as broader facial regions. Conversely, a baseline utilising only EfficientNet tends to focus on more localised patches. The qualitative evidence aligns with our architectural motivation, indicating that the Swin head contributes authentic contextual reasoning rather than simply re-weighting features from the CNN backbone~(see Figure \ref{fig:Results}(c)).

This evaluation is limited to one dataset and one generator family, specifically StyleGAN. Recent studies demonstrate that both CNN and transformer-based detectors may encounter significant performance declines when utilised on unseen datasets or alternative manipulation pipelines~\cite{nadimpalli2022improving}. Our results indicate the potential of Swin-based and hybrid architectures within controlled GAN-based conditions, rather than serving as evidence of complete real-world robustness. Subsequent steps should involve cross-dataset and cross-generator evaluations, as well as the extension of the proposed hybrid design to video deepfakes, which present additional challenges due to temporal consistency and compression artefacts.

\section{Conclusion and Future Scope}
\label{sec:conclusion}

This study examines Swin-Transformer architectures for GAN-based synthetic face detection within defined computational constraints, utilising StyleGAN-generated faces from the 140K Real and Fake Faces dataset. A compact Swin model trained from scratch achieved performance comparable to a robust ELA+CNN baseline. Swin-Tiny, pretrained on ImageNet-1K, yielded a moderate improvement, while the larger Swin-Small variant did not provide additional benefits. The EfficientNet-B0+Swin hybrid demonstrated 99\% accuracy and a 99.44\% recall on a balanced test set of 5,000 images, while being trainable on standard hardware. This suggests that integrating mid-level convolutional features with shifted-window self-attention represents a viable design approach for this binary GAN configuration.

This study concentrates on a specific context: static images, a singular category of GAN-based generators, and a single benchmark dataset. The results should be interpreted as controlled evidence concerning architectural choices, rather than as a comprehensive solution for detecting synthetic media in real-world scenarios. It is necessary to replicate the study across various datasets and generator families, encompassing both diffusion and autoregressive face models. One alternative method involves shifting from images to video, utilising temporal and compression-related cues. A third direction involves clarifying the relationship between model selection, data scale, and energy consumption, with the aim of establishing resource-aware benchmarks for synthetic media forensics.

% arXiv bibliography: use the pre-generated BibTeX output directly.
% This makes the source self-contained and avoids requiring BibTeX during arXiv compilation.

\end{document}